\documentclass[final]{cvpr}

\usepackage{times}
\usepackage{epsfig}
\usepackage{graphicx}
\usepackage{subfigure}
\usepackage{amsmath}
\usepackage{amssymb}
\usepackage{ragged2e}
\usepackage{multirow}
\usepackage{booktabs}

\usepackage[pagebackref=true,breaklinks=true,colorlinks,bookmarks=false]{hyperref}

\renewcommand{\raggedright}{\leftskip=0pt \rightskip=0pt plus 0cm}

\newlength\savewidth\newcommand\shline{\noalign{\global\savewidth\arrayrulewidth
  \global\arrayrulewidth 1pt}\hline\noalign{\global\arrayrulewidth\savewidth}}
  
\newcommand{\mathbbm}[1]{\text{\usefont{U}{bbm}{m}{n}#1}}

\def\cvprPaperID{5839} 
\def\confYear{CVPR 2021}

\begin{document}
\title{\ Learning Dynamic Alignment via Meta-filter for Few-shot Learning}

\author{
Chengming Xu\textsuperscript{\rm 1},
Chen Liu\textsuperscript{\rm 1}, Li Zhang\textsuperscript{\rm 1},\\
Chengjie Wang\textsuperscript{\rm 2},
Jilin Li\textsuperscript{\rm 2},
Feiyue Huang\textsuperscript{\rm 2},
Xiangyang Xue\textsuperscript{\rm 1},
Yanwei Fu\textsuperscript{\rm 1},\\
\textsuperscript{\rm 1}School of Data Science, and MOE Frontiers Center for Brain Science, Fudan University,\\
\textsuperscript{\rm 2}Youtu Lab, Tencent\\
{\tt\small \{cmxu18, chenliu18, lizhangfd, xyxue, yanweifu\}@fudan.edu.cn}\\
{\tt\small \{jasoncjwang, jerolinli, garyhuang\}@tencent.com } \\
}
\maketitle

\begin{abstract}
Few-shot learning (FSL), which aims to recognise new classes by adapting the learned knowledge with extremely limited few-shot (support) examples, remains an important open problem in computer vision. Most of the existing methods for feature alignment in few-shot learning only consider image-level or spatial-level alignment while omitting the channel disparity. Our insight is that these methods would lead to poor adaptation with redundant matching, and leveraging channel-wise adjustment is the key to well adapting the learned knowledge to new classes. Therefore, in this paper, we propose to learn a dynamic alignment, which can effectively highlight both query regions and channels according to different local support information. Specifically, this is achieved by first dynamically sampling the neighbourhood of the feature position conditioned on the input few shot, based on which we further predict a both position-dependent and channel-dependent Dynamic Meta-filter. The filter is used to align the query feature with position-specific and channel-specific knowledge. Moreover, we adopt Neural Ordinary Differential Equation (ODE) to enable a more accurate control of the alignment. In such a sense our model is able to better capture fine-grained semantic context of the few-shot example and thus facilitates dynamical knowledge adaptation for few-shot learning.
The resulting framework establishes the new state-of-the-arts on major few-shot visual recognition benchmarks, including 
\textit{mini}ImageNet and \textit{tiered}ImageNet.
\end{abstract}

 \begin{figure}[t]
     \centering
     \includegraphics[scale=0.45]{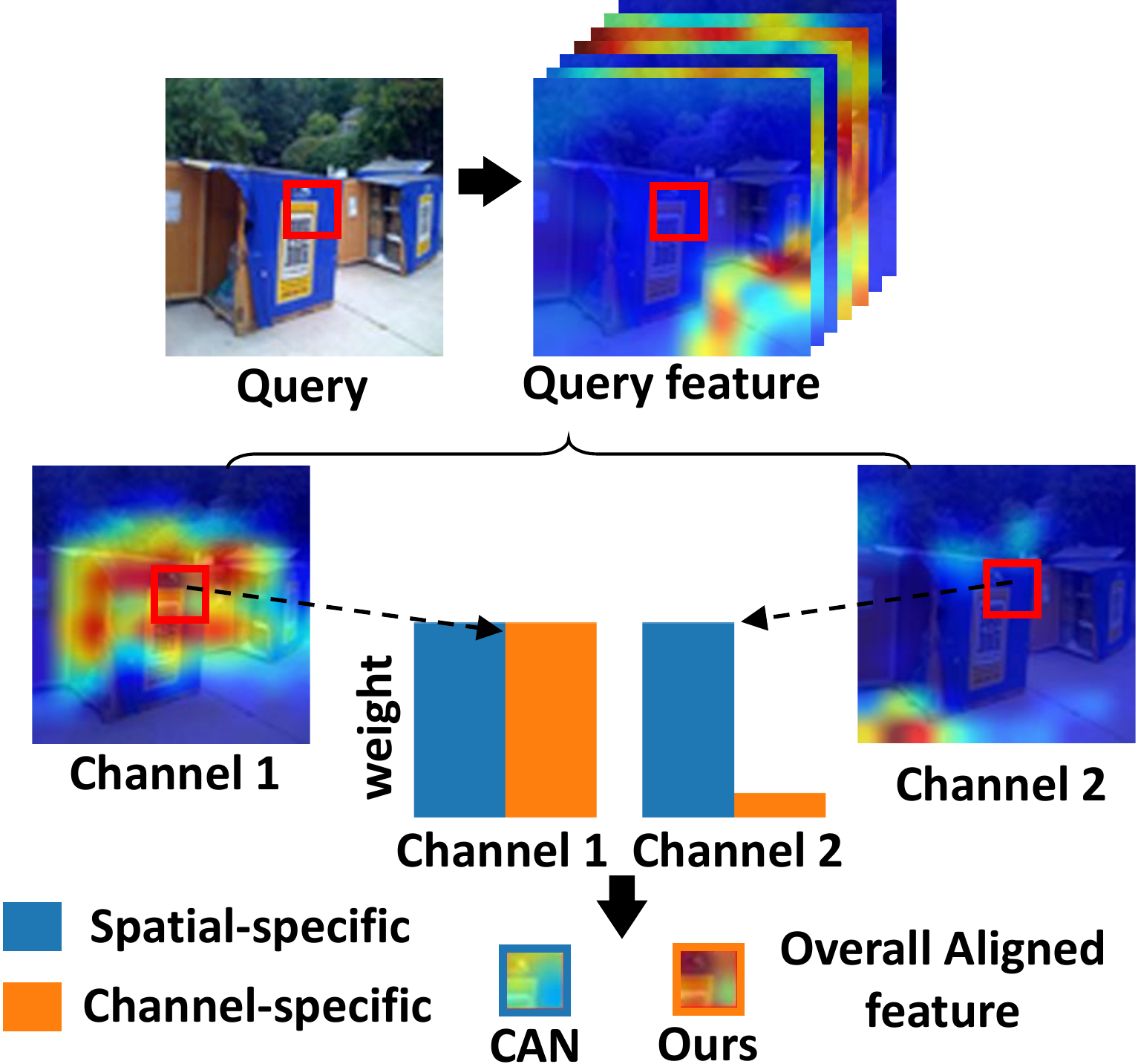}
     \caption{While position-specific alignment cannot eliminate the negative influence of badly-learned channels, channel-specific one can. CAN denotes Cross Attention Network~\cite{hou2019cross}
 which is one of the position-specific alignment models in FSL.}
 \label{fig:intro}
     \vspace{-0.2in}
 \end{figure}

\section{Introduction {\label{sec:intro}}}
Deep learning models have excelled in many computer vision tasks such as image recognition~\cite{he2016deep,simonyan2014very,krizhevsky2012imagenet} and object detection~\cite{he2017mask,xie2019polarmask}.
However, they highly rely on an avalanche of labeled training data and have difficulty transferring the learned knowledge to unseen categories.
For example, an object detector trained on $80$ categories of MSCOCO~\cite{lin2014microsoft} would fail to detect a new class of \textit{mouse}.
This severely limits their scalability to open-ended learning of long tail categories in real-world.
In contrast, learning from extremely constrained (\eg,~one or few) examples is an important ability for humans.
For example,  children have no problem forming the concept of ``giraffe''  by only taking a glance from a picture, or hearing its description as looking like a deer with a long neck.

Motivated by above observations, there has been a recent resurgence of research interest in few-shot learning (FSL)~\cite{koch2015siamese,gidaris2018dynamic,vinyals2016matching,snell2017prototypical,sung2018learning}.
It aims to recognise new classes by adapting the learned knowledge with extremely limited few-shot (support) examples. The most naive baseline for few-shot classification is to
learn a discriminative feature representation by deep convolution neural network. Query features are then assigned with the class label of the nearest support feature. As an alternative to that, learnable metric like RelationNet~\cite{sung2018learning} is proposed, where a binary classifier consisting of multiple layers of neural networks is utilised to calculate the similarity between two images. In such a framework, there is no exploration of more valuable information for specific query image when being classified with different support images, which leads to bad generalization ability. Although there are some recent works targeting feature alignment for FSL, such as CAN~\cite{hou2019cross} and FEAT~\cite{ye2020feat}, which aggregates support knowledge to formalize an alignment function for query samples, we claim that these methods mainly have the following drawbacks. 
(1) \textit{Roughness.} Due to limited knowledge in FSL, every channel would contribute to the prediction when comparing support feature to the query feature. Moreover, since the extracted features only have low spatial resolution, the information discrepancy among spatial positions is far smaller than that among channels for each query features. However, methods like CAN and FEAT do not focus on channel-level information. For example as shown in Fig.~\ref{fig:intro}, while the spatial-specific alignment can set a large weight to the red box inside the target object, such a large weight is assigned to both well-learned and badly-learned channels. This would result in low signal in this region in the overall feature after alignment. In contrast, channel-specific alignment depends on the quality of each channel, thus being able to set low weight to channel 2, resulting in a better overall feature.
(2) \textit{Redundant matching.} There exists lots of redundancy in the support knowledge. For example, when classifying one position containing the target object, if there are several regions that also have this object, then comparing to one of them is enough for classification. Nonetheless, the existing methods utilize the whole support feature when aligning each query position, which is inefficient. 
(3) \textit{Inflexible alignment.} The alignment strategy in these works only runs one time for all tasks. Thus, for those difficult ones, the alignment may be insufficient to appropriately embed the support knowledge into query feature. 

Therefore to solve these problems, in this paper we propose a novel dynamic feature alignment strategy. In detail, we turn to dynamic filters~\cite{jia2016dynamic, zhang2020dynamic} for tackling this problem. We first predict a dynamic meta-filter with both position-specific and channel-specific filter weights based on small neighbourhood of each position. This filter can contain adequate information to inform the model of the most important regions and channels that need to be highlighted. Hence, applying this filter to align the query features will culminate in more effective representations for recognition. Meanwhile, using the neighborhood rather than the whole support feature directly decreases the number of knowledge source and the redundancy. To alleviate the problem resulted from fixed neighbor, we adopt a dynamic sampling strategy that all of the feature positions can be selected via learning an offset conditioned on the input few shot pairs. 
Intuitively, this learned sampling allows the network to better capture position based semantic context of the few-shot example. To further make the alignment more adaptive to harder tasks, a direct way is to recursively apply the support knowledge to query feature through repetitive alignment. However, it is hard to control the extent of alignment by fixed hyper-parameter for various tasks. Consequently we modify the direct intuition of recursive alignment into an adaptive manner by using Neural Ordinary Differential Equation (ODE) which makes continuous the residual alignment procedure and takes an adaptive step size to get the final solution for the corresponding ODE. After achieving the aligned query feature, we utilize a meta-classifier to get the final prediction where the support knowledge is aggregated by unlearnable operations to form a classifier, which can avoid the adaptation problem that learnable classifiers are fully dependent on meta-train set and cannot adjust well for data in novel categories.

The contributions of this work are as follows:
\vspace{-0.05in}
\begin{enumerate}
    \item We propose to learn a novel dynamic meta-filter for more effective and efficient feature alignment in FSL. 
    \item We introduce dynamic sampling and grouping strategy to further improve the flexibility and efficiency of basic dynamic meta-filter. 
    \item Neural ODE is, for the first time, leveraged to help model get better representation for FSL.
\end{enumerate}

\section{Related work}
\noindent\textbf{Few-shot recognition.} Few-shot learning (FSL) is a surging research topic which aims to learn patterns with a set of data (base classes) and adapt to a disjoint set (new classes) with limited training data. Few-shot image classification is the one with most focus and researches. There are two main ways to tackle this problem. One is optimization-based methods~\cite{ravi2016optimization, finn2017model, nichol2018first, li2017meta, sun2019meta}, which firstly train a network with base class data, then finetune the classifier or the whole network with support data from unseen classes.

The metric-based method, on the other hand, is designed to solve FSL by applying an existing or learned metric on the extracted features of images. MatchingNet~\cite{vinyals2016matching} adopts memory module to merge the information in each task and cosine distance as the metric to classify unseen data. ProtoNet~\cite{snell2017prototypical} proposes the prototype as a simple representation of each category and adopts euclidean distance as the metric. RelationNet~\cite{sung2018learning} uses network to learn the relation, which is taken as similarity, between input few-shot pairs. CAN~\cite{hou2019cross}, based on RelationNet, learns an attention module to highlight the correct region of interest to help better classification. 
The key dissimilarity between our model and the existing metric-based methods is that we propose to dynamically sample local context and based on that to learn position and channel-dependent relationship which has never been explored before.

\noindent\textbf{Dynamic Sampling.}
Sampling local context in a proper manner has been studied in the literature for a long time. 
Prior to the rise of deep learning, SIFT~\cite{lowe1999object} and DPM~\cite{felzenszwalb2009object} attempt to construct a good adaptive kernel. 
In the era of deep learning, convolutional kernel samples the neighbourhood of corresponding position in a uniform manner. For instance, a uniform $9$-neighbourhood is sampled for a $3\times3$ kernel. 
Though this strategy paves the way for the success of the convolutional neural networks (CNNs), it is still limited in capturing position based semantic context. Our method is related to deformable convolution~\cite{dai2017deformable}, where, instead of uniform sampling, a specific region is learned for each position. A fundamental difference between our model and the existing ones is that they only learn the offset dependent on the input feature while the filter weights are fixed for all inputs. In contrast, our model learns the offset to sample the neighbourhood of the feature position. The position and channel dependent filter is then predicted based on the sampled neighbourhood. Intuitively, this learned sampling allows the network to better capture position based semantic context of the few-shot example.

\noindent\textbf{Input dependent weights.}
The basic idea of input dependent weights is to control the model weight not by directly optimizing but another learnable module. 
~\cite{jia2016dynamic} developed an idea of ``dynamic conolution'', that is predicting a dynamic convolutional filter for each feature position.
HyperNet~\cite{ha2016hypernetworks} builds additional module to generate input-dependent weight for a RNN. ~\cite{bertinetto2016learning} firstly imports this idea to FSL.  ~\cite{gidaris2018dynamic} proposes to generate final classifier weights based on pretrained classifier weights and the input information during test phase. 
~\cite{lifchitz2019dense} implants several new filters to the top layer while freezes other part of the model during test phase. 
The newly added weights is finetuned using support data. ~\cite{qiao2018few} aims to bridge the activation and classifier weights via additional sub-network.
Unlike the above works that generate input-dependent or category-dependent weights,
we focus on the local information and discrepancy between different positions and channels of query images. Our model can generate both position and channel specific filter weights which are especially beneficial for us to model the context of each feature position for target class, thus leading to more effective few-shot classification.


\noindent\textbf{Neural Ordinary Differential Equation.} Neural ODE was proposed by Chen et. al. in \cite{chen2018ode}, which treats forward pass of a residual network as a discrete form of ODE. Under such condition, neural networks can be modified into neural ODEs where a time variable is introduced to control the output. There is a line of works targeting more efficient and robust neural ODE~\cite{massaroli2020dissecting, rubanova2019latent, dupont2019augmented}, and another line is the application of neural ODE to other tasks. For example, ODE$^2$VAE~\cite{yildiz2019ode2vae} utilizes neural ODE for modeling  trajectory of high-dimensional data, and Vid-ODE~\cite{park2020vid} uses it in video generation. 
In this paper, neural ODE is, for the first time, introduced to few-shot learning to help learn a dynamic alignment between support and query features.

\begin{figure*}[t]
    \centering
    \includegraphics[scale=0.6]{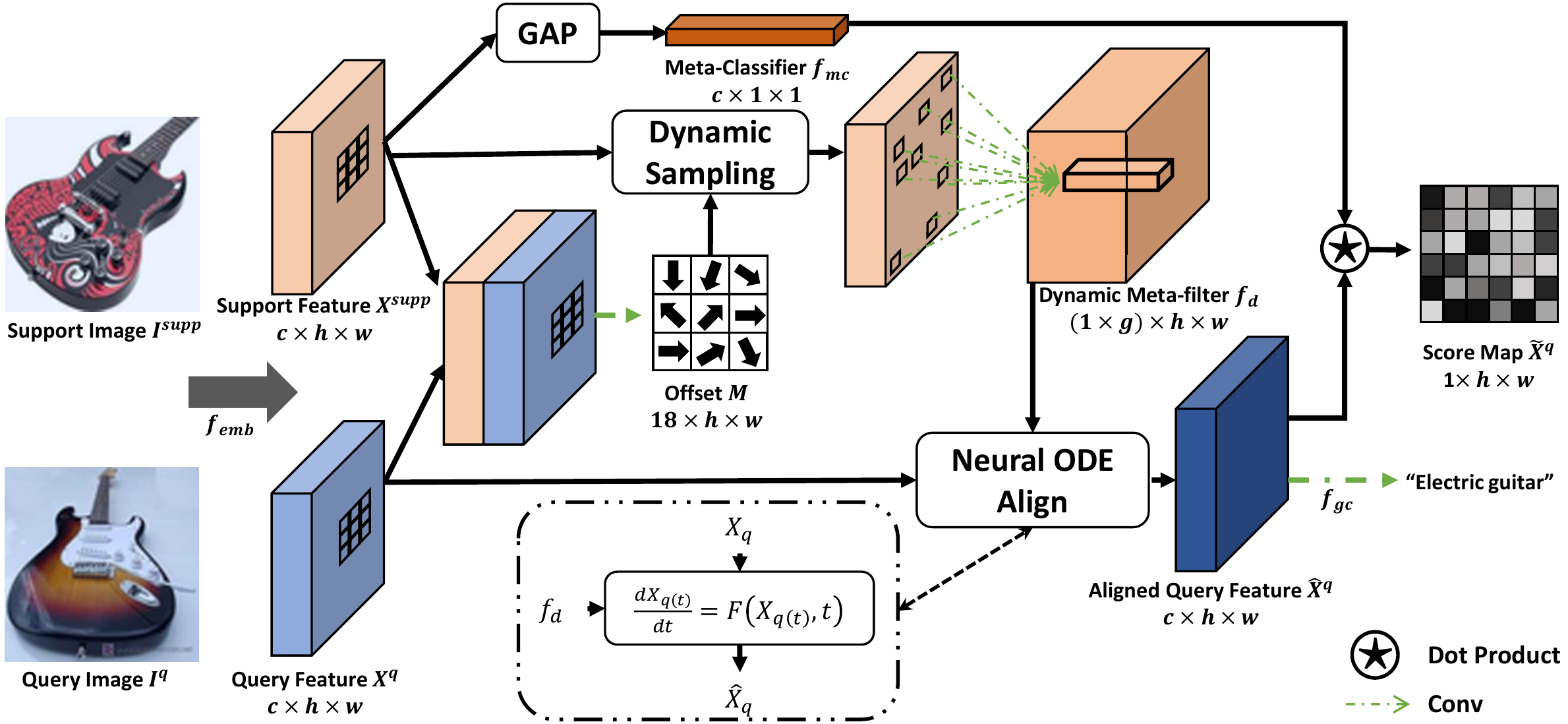}
    \caption{Schematic illustration of our proposed model for 1-shot task. Support and query features are extracted with a backbone network $f_{emb}$. Then two features are employed to predict an offset map using the local knowledge, which is used to dynamically sample the support feature to collect useful information for generating dynamic meta-filter for each query position. The filter is applied to align the query feature which is classified with a meta-classifier. GAP denotes global average pooling.}
    \label{fig:model}
    \vspace{-0.15in}
\end{figure*}

\section{Problem formulation}
The assumption of data split in FSL is different from common supervised learning. Suppose that we have two sets of data, meta-train set $\mathcal{D}_s=\left\{ \left(\mathbf{I}_{i},y_{i}\right),y_{i}\in\mathcal{C}_{s}\right\}$ and meta-test set $\mathcal{D}_t=\left\{ \left(\mathbf{I}_{i},y_{i}\right),y_{i}\in\mathcal{C}_{t}\right\}$ standing for collections of base class data and novel class data. $\mathcal{C}_s$ and $\mathcal{C}_t$ represent base and novel category sets respectively ($\mathcal{C}_s \cap \mathcal{C}_t = \emptyset $). The goal of FSL is to train a model on $\mathcal{D}_s$ which is well generalized to $\mathcal{D}_t$. According to the commonly-used setting, we can access to few (\eg, one or five) labelled data from each category of $\mathcal{C}_t$. These ground truth provide the supervision we can use to transfer the knowledge learned from base classes to novel classes. 

We follow former methods \cite{snell2017prototypical,sung2018learning} to adopt an $N$-way $K$-shot meta-learning strategy. Here $N$ denotes number of categories in one episode and $K$ stands for number of samples for each category in support set. 
In detail, we sample $N$ categories from $\mathcal{C}_s$ for training and $\mathcal{C}_t$ for testing, $K$ instances each for these selected categories to construct a support set $\mathcal{S}=\left\{ \left(\mathbf{I}_{i}^{supp},y_{i}^{supp}\right)\right\}$. 
Similarly we sample $Q$ pieces of data from the $N$ categories as query set $\mathcal{Q}=\left\{ \left(\mathbf{I}_{i}^q,y_{i}^q\right)\right\}$, and $\mathcal{S} \cap \mathcal{Q} = \emptyset$.

\section{Dynamic Alignment Network}
Our model is illustrated in Fig.~\ref{fig:model} where four sub-modules are used: feature extractor $f_{emb}$, dynamic alignment module $f_d$, meta-classifier $f_{mc}$ and a global-classifier $f_{gc}$. Given each pair of support and query image $(\mathbf{I}^{supp}, ~\mathbf{I}^q)$, first we use $f_{emb}$ to extract feature maps $X^{supp}=f_{emb}(\mathbf{I}^{supp}), ~X^q=f_{emb}(\mathbf{I}^q)$. The size of each feature map is $c\times h\times w$, where $c, h, w$ indicates the number of channel, height and width, individually. The feature for each category can be represented as the average of all support feature maps in this class. For simplicity we will still use $X^{supp}$ to stand for feature maps of each category. Next the dynamic alignment module $f_d$, which consists of a dynamic meta-filter and adaptive alignment, is used to align query feature with both position and channel-specific support knowledge. Finally, a confidence score of similarity between the adjusted query feature and corresponding support feature is provided by the meta-classifier $f_{mc}$. Meanwhile, a global classifier $f_{gc}$ is also applied to the query feature to get a prediction of real category. All these modules are trained with a few-shot classification loss and a global classification loss. In the following context we describe our dynamic meta-filter by first explaining the na\"ive alignment model in Sec.~\ref{sec:naive} and then proposing and improving our dynamic meta-filter from multiple points in Sec.~\ref{sec:dmf}. Then we complete the framework by introducing the meta-classifier (Sec.~\ref{sec:mc}) and some implementation details (Sec.~\ref{sec:detail})

\subsection{Na\"ive feature alignment \label{sec:naive}}
After extracting support and query features, several former studies show that applying an alignment or transformation to the query features based on support knowledge can help the model filter out feature part most beneficial to classification, thus improving the performance. A naive alignment can be written in the following form:
\begin{equation}
    \hat{X}^q = g(X^q, A(X^q, \mathcal{X}))
\end{equation}
where $g$ is the aligning operation, $A$ is alignment basis and $\mathcal{X}$ is knowledge source for generating $A$. For CAN~\cite{hou2019cross}, $g$ is multiplication, $\mathcal{X}$ is single support feature and $A$ is cross attention between two feature maps. For FEAT~\cite{ye2020feat}, $g$ is summation, $\mathcal{X}$ is whole support set and $A$ is weighted average of $\mathcal{X}$ with a self-attention module. The common ground of these two methods lies in two aspects: (1) Both methods omit channel-level alignment. CAN focuses on spatial-level attention while FEAT uses an instance-level one. (2) All available support knowledge is used. These two properties lead to the weaknesses including roughness, redundant matching and inflexible alignment as discussed in Sec.~\ref{sec:intro}. To this end, we propose a new alignment framework in the form of dynamic meta-filter (DMF). 

\subsection{Dynamic Meta-filter for Adaptive Alignment \label{sec:dmf}}
Concretely, given the support and query feature pair $\{X^{supp}, ~X^q\}$, instead of directly generating a filter from support features, we first apply a convolution layer $f_{\psi}$ parameterized by $\psi$ with kernel size 3 to the support feature $X^{supp}$, which outputs a tensor $\psi(X^{supp})\in \mathbb{R}^{(c\times k\times k)\times h\times w}$ where $k$ denotes the kernel size of DMF. Then, for each position $(i,j)$ we can have a tensor of size $c\times k\times k$:
\begin{equation}
    f_d(i, j)=\sigma(\psi *\mathcal{B}_3(X^{supp}_{:,i,j}))
    \label{eq:2}
\end{equation}
where $\mathcal{B}_3$ means a $3$ width neighbor, $*$ denotes convolution, $\sigma$ is the Sigmoid function used to control the scale.
This $f_d(i,j)$ can be seen as $c$ convolution filters with kernel size $k$. Convolving $f_d(i,j)$ to the corresponding position of $X^q$ with $c$ groups leads to a refined query feature $\hat{X_{q}} \in \mathbb{R}^{c\times h\times w}$, where 
\begin{equation}
    \hat{X}^{q}_{:,i,j}=X^{q}_{:,i,j}+f_d(i,j)*_c\mathcal{B}_k(X^{q}_{:,i,j})
     \label{eq:3}
\end{equation} 
where $*_c$ denotes convolution with group $c$. In this way, each channel is rescaled by a weight computed by our DMF and the query feature is thus aligned according to the information collected from support feature in a both position-varying and channel-varying manner. Meanwhile, the source of support information for each query position is directly decreased from whole support feature to a small grid.

The dynamic meta-filter described above is both efficient and capable of dealing with fine-grained support knowledge, but drawbacks are also obvious: (1) The convolutional kernel used to generate DMF can only have a fixed kernel size, which would limit the receptive field, increasing the probability of failure to collect useful support information. (2) The DMF would be large and time consuming if the feature map has large dimension.  (3) Despite a dynamic filter, the alignment is fixed, thus not flexible. 
Hence, we further propose three improvements to solve these problems.

\noindent\textbf{Dynamic Sampling}
To help the model be more flexible to fetch any required features, we utilize a dynamic sampling strategy. In detail, the fixed convolution region in $f_{\psi}$ is replaced with a learnable one. Conditioned on the input support and query feature pair, a neighbourhood map $M\in\mathbb{R}^{9\times h\times w}$ of each feature position is predicted with a convolution layer $f_{\eta}$ of kernel size 5:
\begin{equation}
    M = f_{\eta}(X^{supp} \| X^q)
\end{equation}
where $\cdot\|\cdot$ denotes concatenation. Each row of the neighbourhood map $M_{:, i, j}\in\mathbb{R}^9$ denotes the regions containing useful information for generating the filter at $(i,j)$. This map is then used to dynamically sample knowledge from $X^{supp}$ for each position to generate the DMF:
\begin{align}
    \tilde{\mathcal{B}}_3(X^{supp}_{:,i,j}) &= \mathrm{Sample}(X^{supp}_{:,i,j}, M_{:,i,j}) \\
    f_d(i, j) &=\sigma(\psi * \tilde{\mathcal{B}}_3(X^{supp}_{:,i,j}))
\end{align}
Through dynamic sampling, we can increase the probability of exploring the useful positions while keeping a low volume of source knowledge. Among the many ways to implement the dynamic sampling, we imitate the deformable convolution (DeformConv)~\cite{dai2017deformable} where the neighborhood map is realized in the form of horizontal and vertical offsets $M\in\mathbb{R}^{18\times h\times w}$ and sampling is based on the offset from each position. Note that \cite{wu2019parn} also adopts DeformConv in few-shot learning. However, they directly employ DeformConv to both the support and query features and map them into another latent space, which are then used to measure the correlation. No information interaction is conducted between support and query data. Compared to this method, the dynamic sampling in our model is predicted based on both support and query knowledge and used to generate a dynamic position and channel-dependent meta-filter which is then utilized to embed the knowledge from support images into query features.

\noindent\textbf{Grouping Strategy \label{sec:group}}
To decrease the computation cost resulted from large number of filters, we make a simple assumption: similar information is shared among some groups of channels in feature maps. Attributed to that, we follow the existing works on group convolution~\cite{xie2017aggregated} to gather these channels together to share one filter. 
Specifically, we equally segment the query feature map into $g$ groups. The output channel of the meta-filter generator $f_{\psi}$ is then modified from $c\times k\times k$ to $g\times k\times k$. Due to grouping the model only needs to generate a $g$-channel filter for each position, thus more lightweight. It is noteworthy that when $g=1$, our DMF works in the same way as CAN and FEAT where all channels of each position are processed with the same weight. Such a strategy is shown to be less powerful than a larger $g$ in our experiments, which proves our advantage against the other methods.

\noindent\textbf{Adaptive Alignment} The most direct solution for more difficult tasks is to recur the alignment process to get a more refined feature, such as in \cite{wu2017recursive} where a learned spatial transformation is repeatedly used on the image features. By this means, the query feature can be written as:
\begin{align}
    X^q_{t+1}&=X^q_t+F(X^q_t, f_d),\quad t=0,\cdots,T-1 \\
    \hat{X}^q &= X^q_{T}
    \label{eq:residual}
\end{align}
where $F$ is the dynamic convolution. Such a method raises another problem --- tuning the extra hyper-parameter of recursion depth $T$. In \cite{wu2017recursive} the authors tried several choices of this hyper-parameter for one dataset. However, for different tasks, the depth variable should be varying depending on the level of complication. Therefore for FSL, it is inappropriate to use a fixed recursion depth on episodes containing various categories. As an alternative, we refer to the Neural ODE~\cite{chen2018ode}. By gradually decreasing the time step $t$, the recursive residual equation~\ref{eq:residual} can be transformed from an Euler discretization form into an ODE:
\begin{equation}
    \frac{dX^q(t)}{dt}=F(X^q(t), t)
    \label{eq:ode}
\end{equation}
Then $\hat{X}^q$ is set as the solution of Eq.~\ref{eq:ode} given initial condition $X^q$. Such a form is better for two reasons. First, using modern ODE solvers such as Dormand-Prince method can get more accurate solution than Euler's method. Second, by taking adaptive step size, the recursive depth is thus data-dependent, thus getting rid of hyper-parameter $T$.

\subsection{Meta classifier \label{sec:mc}}
After we achieve a dynamically aligned query feature $\hat{X}^q$, we follow the CAN~\cite{hou2019cross} to take a simple way to form the classifier which does not need any learnable parameters. Formally, we aggregate $X^{supp}$ by global average pooling into $\bar{X}^{supp}\in \mathbb{R}^{c\times 1\times 1}$, which contains the global information of $\mathbf{I}^{supp}$. We then use it as the weights for a convolutional filter, named meta-classifier (MC) $f_{mc}$, with $c$ as the number of channel and $1\times 1$ as kernel size. 
Applying this filter to query feature leads to a tensor $\tilde{X}^q\in \mathbb{R}^{1\times h\times w}$. 
Through the calculation we directly compare the global support feature with local query feature on each corresponding channel. Therefore, if $\mathbf{I}^q$ has the same category as $\mathbf{I}^{supp}$, $\tilde{X}^q$ should have high value in most positions. 

\subsection{Implementation details \label{sec:detail}} 
\noindent\textbf{Objective function.} Our loss function is defined in the same way as in CAN. For each input few-shot pair with all inputs and outputs $\{X^{supp}, y^{supp}, X^{q}, y^{q}, \hat{X}^{q}, \tilde{X}^{q} \}$, denote $\mathcal{X}$ as the set of results from $f_{mc}$ between $\mathbf{I}^q$ and all support features, the objective function can be written as follow:
\begin{align}
    \mathcal{L}&=\ell_{f}+\frac{1}{2} \ell_{g} \\
     \ell_{g} &= -(\log \textrm{softmax}(f_{gc}(\hat{X}^q)))^Ty^{q} \\
    \ell_{f} &= \frac{1}{hw}\sum_{s,t}\log\frac{e^{-\tilde{X}^q_{s,t}}}{\sum_{\tilde{X}\in\mathcal{X}}(1+e^{-\tilde{X}_{s,t}})}\cdot \mathbbm{1}_{y^q=y^{supp}}
\end{align}
where $\ell_{f}$ is for the few-shot $N$-way classification, and $\ell_{g}$ is for the global many-shot $|\mathcal{C}_s|$-way (\textit{e.g.}, 64 for \textit{mini}ImageNet) classification.

\noindent\textbf{Network Structure.} We use the same ResNet12 as in \cite{ye2020feat} for extracting image features. To enlarge the use of position-specific property of our DMF, we remove the last pooling layer of the backbone so that the spatial size of output feature map is doubled to $11 \times 11$. Note that no extra parameters or capacity are introduced here. The convolution between our DMF and query features is implemented by first unfolding the query features along the spatial dimension and calculate the dot product between the unrolled features and DMF. The whole operation is highly efficient which will be shown in supplementary material.

\begin{table*}[h]
 \centering
 {
  \begin{tabular}{ l|l|cccc}
  \hline
 \multirow{2}{*}{Model} & \multirow{2}{*}{Backbone} &  \multicolumn{2}{c}{\textit{mini}ImageNet}  &   \multicolumn{2}{c}{\textit{tiered}ImageNet} \tabularnewline
 \cline{3-6}
& &  1-shot  &   5-shot &   1-shot & 5-shot\\

   \shline
   ProtoNet~\cite{snell2017prototypical} & \multirow{5}{*}{Conv4} & 49.42$\pm$0.78 & 68.20$\pm$0.72 & 53.31$\pm$0.89 & 72.69$\pm$0.74\tabularnewline
   \cline{3-6}
   MatchingNet~\cite{vinyals2016matching}&  & 43.56$\pm$0.84 & 55.31$\pm$0.73 & --- & ---\tabularnewline
   \cline{3-6}
   RelationNet~\cite{sung2018learning} &  & 50.44$\pm$0.82 & 65.32$\pm$0.70 & 54.48$\pm$0.93 & 71.32$\pm$0.78\tabularnewline
   \cline{3-6}
   MAML~\cite{finn2017model} &  & 48.70$\pm$1.75 & 63.11$\pm$0.92 & --- & ---\tabularnewline
   \cline{3-6}
   Dynamic Few-shot~\cite{gidaris2018dynamic} &  & 56.20$\pm$0.86 & 72.81$\pm$0.62 & --- & ---\tabularnewline
    \hline
     LEO~\cite{rusu2018meta} & \multirow{6}{*}{WRN-28} & 61.76$\pm$0.08 & 77.59$\pm$0.12 & 66.33$\pm$0.05 & 81.44$\pm$0.09\tabularnewline
     \cline{3-6}
     PPA~\cite{qiao2018few} &  & 59.60$\pm$0.41 & 73.74$\pm$0.19 & --- & ---\tabularnewline
     \cline{3-6}
     Robust dist++~\cite{dvornik2019diversity} &  & 63.28$\pm$0.62 & 81.17$\pm$0.43 & --- & ---\tabularnewline
     \cline{3-6}
     wDAE~\cite{gidaris2019generating} &  & 61.07$\pm$0.15 & 76.75$\pm$0.11 & 68.18$\pm$0.16 & 83.09$\pm$0.12\tabularnewline
     \cline{3-6}
     CC+rot~\cite{gidaris2019boosting} &  & 62.93$\pm$0.45 & 79.87$\pm$0.33 & 70.53$\pm$0.51 & 84.98$\pm$0.36\tabularnewline\cline{3-6}
    FEAT~\cite{ye2020feat} &  & 65.10$\pm$0.20 & 81.11$\pm$0.14 & 70.41$\pm$0.23 & 84.38$\pm$0.16\tabularnewline
    \hline
     TapNet~\cite{yoon2019tapnet} & \multirow{13}{*}{Res-12} & 61.65$\pm$0.15 & 76.36$\pm$0.10 & --- & ---\tabularnewline
     \cline{3-6}
     SNAIL~\cite{mishra2017simple} &  & 55.71$\pm$0.99 & 68.88$\pm$0.92 & --- & ---\tabularnewline
     \cline{3-6}
     MetaOptNet~\cite{lee2019meta} &  & 62.64$\pm$0.61 & 78.63$\pm$0.46 & 65.99$\pm$0.72 & 81.56$\pm$0.53\tabularnewline
     \cline{3-6}
     TADAM~\cite{oreshkin2018tadam} &  & 58.50$\pm$0.30 & 76.70$\pm$0.30 & --- & ---\tabularnewline
     \cline{3-6}
     DC~\cite{lifchitz2019dense} &  & 62.53$\pm$0.19 & 78.95$\pm$0.13 & --- & ---\tabularnewline
     \cline{3-6}
     \cline{3-6}
     VFSL~\cite{fei2020meta} &  & 61.23$\pm$0.26 & 77.69$\pm$0.17 & --- & ---\tabularnewline
     \cline{3-6}
     CAN~\cite{hou2019cross} &  & 63.85$\pm$0.48 & 79.44$\pm$0.34 & 69.89$\pm$0.51 & 84.23$\pm$0.37\tabularnewline
     \cline{3-6}
      FEAT~\cite{ye2020feat} &  & 66.78$\pm$0.20 & 82.05$\pm$0.14 & 70.80$\pm$0.23 & 84.79$\pm$0.16\tabularnewline \cline{3-6}
      DeepEMD~\cite{zhang2020deepemd} &  & 65.91$\pm$0.82 & 82.41$\pm$0.56 & 71.16$\pm$0.87 & \textbf{86.03}$\pm$0.58\tabularnewline \cline{3-6}
      E$^3$BM~\cite{liu2020ensemble} &  & 63.80$\pm$0.40 & 80.10$\pm$0.30 & 71.20$\pm$0.40 & 85.30$\pm$0.30\tabularnewline \cline{3-6}
      DSN-MR~\cite{simon2020adaptive} &  & 64.60$\pm$0.72 & 79.51$\pm$0.50 & 67.39$\pm$0.82 & 82.85$\pm$0.56\tabularnewline \cline{3-6}
      Net-Cosine~\cite{liu2020negative} &  & 63.85$\pm$0.81 & 81.57$\pm$0.56 & --- & ---\tabularnewline\cline{1-6}
     Ours & Res-12 &\bf 67.76$\pm$0.46 & \bf 82.71$\pm$0.31 & \bf 71.89$\pm$0.52 & 85.96$\pm$0.35 \tabularnewline
     \hline
  \end{tabular}
 }
\caption{\label{tab:result}5-way few-shot accuracies with $95\%$ confidence interval on \textit{mini}ImageNet
and \textit{tiered}ImageNet.}
 \vspace{-0.15in}
\end{table*}

\section{Experiments}
\subsection{Datasets and setting}
\noindent\textbf{Datasets.}  Our experiments are conducted on two datasets. \textit{mini}ImageNet dataset~\cite{vinyals2016matching}, containing 600 images with each of the 100 categories, is a small subset of ImageNet. We follow the split in \cite{ravi2016optimization}, where 64, 16, 20 classes are used for train,  validation and test, respectively.  \textit{tiered}ImageNet dataset~\cite{ren2018meta} is a larger subset of ILSVRC-12 dataset. It consists of 34 categories with 779,165 images in total. These categories are further broken into 608 classes, where 351 classes are used for training, 97 for validation and 160 for testing. The size of images in \textit{mini}ImageNet is $84\times 84$ and in \textit{tiered}ImageNet, $224\times 224$. Images in \textit{tiered}ImageNet are resized to $84\times 84$ before training and testing. 

\noindent\textbf{Experimental setup.} We empirically set the groups $g$ as 64 for 1-shot tasks on both datasets, 160 for 5-shot \textit{mini}ImageNet and 320 for 5-shot \textit{tiered}ImageNet. Kernel size $k$ is assigned as 1 for all tasks. Stochastic Gradient Descent (SGD)~\cite{bottou2010large} with $5e-4$ weight decay is used to optimize our model. For \textit{mini}ImageNet, the initial learning rate is set as $0.35$ and $0.05$ for \textit{tiered}ImageNet. Cosine learning rate decay \cite{loshchilov2016sgdr} is used along with SGD. Random cropping, horizontal flipping, color jittering and random erasing~\cite{zhong2017random} are adopted for data augmentation during training, which is the same as in CAN~\cite{hou2019cross}.
We test 2000 episodes sampled from meta-test set for all experiments.

\noindent\textbf{Evaluation benchmark.} We report the accuracy and $95\%$ confidence interval (CI) of 5-way 1-shot and 5-way 5-shot settings when comparing with the existing methods. For ablation study, only accuracy is reported. 

\noindent\textbf{Competitors.} To show the efficacy of our model, we compare it with several previous methods for example Prototypical Network (ProtoNet)~\cite{snell2017prototypical}, RelationNet~\cite{sung2018learning}, MetaOptNet~\cite{lee2019meta}, Cross Attention Network (CAN)~\cite{hou2019cross}, etc. These models are chosen because they are among the best few-shot learning models and also the results with the same setting have been reported in the original paper.

\subsection{Comparison with state-of-the-art}
We compare our model with the competitors in Tab.~\ref{tab:result}, where the accuracies and 95\% CI of 5-way 1-shot and 5-way 5-shot tasks on two datasets are shown. Note that different backbone structures are used among these models.

\noindent\textbf{\textit{mini}ImageNet result.} As shown in the Tab.~\ref{tab:result}, on \textit{mini}ImageNet, all of the models with Conv4 are worse than ours, which in part results from the low capacity of the backbone. When comparing with models trained with the same backbone, the 1-shot accuracy of our model is still 0.98\% higher than the best one, \textit{i.e.}, FEAT (66.78\%). When comparing with models using WRN-28-10 which has larger capacity than ResNet12, our model is still outstanding. On 5-shot task, our model is 0.30\% higher than DeepEMD which is the strongest competitor using ResNet12 backbone, and 1.54\% higher than Robust dist++ which is the best 5-shot model with WRN backbone. It reflects that the ability of our model to deal with extremely limited data is better than most of the competitors, while ours is well-designed to draw the correct characteristics in common among limited images. Moreover, it is worth noting that the additional parameters of our model to the backbone are just two convolutional layers which is quite lightweight due to the group strategy. In contrast, most of these competitors, for example, TADAM and E$^3$BM, have much more parameters. Also, some of the state-of-the-art methods for example FEAT and DeepEMD adopt pre-train to get a good initialization of the backbone for the meta-training stage so that their proposed modules used for few-shot recognition can be trained well. Compared with them, our proposed Dynamic Meta-filter does not depend on initialization, thus  pretrain-free. In this way, our model can be trained with an ensemble of global classification loss and few-shot classification loss, which would be faster than those competitors.

\noindent\textbf{\textit{tiered}ImageNet result.} 
Results in Tab.~\ref{tab:result} show that on \textit{tiered}ImageNet we have 0.69\% accuracy boost over the best competitor on 1-shot classification, which also proves the above conclusion. For 5-shot tasks, our model is 0.07\% worse than DeepEMD. One possible reason is that \textit{tiered}ImageNet is a larger dataset with more base categories, hence the backbone itself can already learn a good representation with 5 support samples for each novel class. Moreover, DeepEMD receives marginally better results at the cost of massive training and inference time resulted from a more complicated classifier involving Quadratic Programming. In contrast, our framework is more succinct.

\subsection{Model analysis}
To further validate the effectiveness of our method, we conduct a series of ablation studies on \textit{mini}ImageNet. We first show some experimental proofs of the derivation of our method, then we show some other results on the hyper-parameters. The results are shown in Tab.~\ref{tab:ablation}.

\begin{table}[htb]
\begin{minipage}{0.22\textwidth} 
 \centering 
{
    {
  \begin{tabular}{ l|cc}
  \hline
DS & K=1 & K=5 \\
\shline
\checkmark &  \bf 67.76  & \bf 82.71  \\
$\times$    &  66.73 & 81.20\\
\hline
\end{tabular}}
}
\end{minipage}
\begin{minipage}{0.22\textwidth} 
 \centering 
{
{
  \begin{tabular}{l|cc}
  \hline
Pool & K=1 & K=5 \\
\shline

\checkmark   & 67.12 &  81.54 \\
$\times$  &\bf 67.76  &\bf  82.71 \\
\hline
\end{tabular}}}
\end{minipage}
\vspace{0.1in}
\\
\begin{minipage}{0.22\textwidth} 
\centering(a)
\end{minipage}
\begin{minipage}{0.22\textwidth}
\centering(b)
\end{minipage}
\vspace{0.1in}
\\
\begin{minipage}{0.22\textwidth} 
\centering 
{
  \begin{tabular}{ l|cc}
  \hline
Group & K=1 & K=5 \\
\shline

1 & 66.48 & 81.40 \\
64 & \bf 67.76 & 82.17\\
160 & 67.61 &\bf 82.71\\
320 & 67.51 &  82.71\\
640 & 67.43 & 82.34\\
\hline
\end{tabular}}
\end{minipage}
\begin{minipage}{0.2\textwidth} 
\centering 
{
  \begin{tabular}{ l|cc}
  \hline
Align & K=1 & K=5 \\
\shline

0 & 64.20 & 80.24\\
1 & 67.26 & 82.14\\
2 & 67.27 & 81.81\\
3 & 67.45 & 81.19\\
ODE &\bf 67.76  &\bf  82.71\\
\hline
\end{tabular}}
\end{minipage}
\vspace{0.1in}
\\
\begin{minipage}{0.22\textwidth} 
\centering(c)
\end{minipage}
\begin{minipage}{0.22\textwidth}
\centering(d)
\end{minipage}
\\
\vspace{-0.1in}
\caption{\label{tab:ablation} Ablation Studies on \textit{mini}ImageNet 5-way tasks. We show 1-shot(K=1) and 5-shot(K=5) results. (a) \textbf{Dynamic Sampling}: Full model is compare with one without dynamic sample. (b) \textbf{Pooling}: we compare model trained with and without the last pooling layer in the Res-12 backbone. (c) \textbf{Groups}: we compare the number of groups in dynamic conv. (d) \textbf{Instantiation}: we try different structures based on our method, including models with no alignment, fixed number of alignment and ODE alignment.}
 \vspace{-0.2in}
\end{table}

\begin{figure}[htb]
 \def \imheight {0.07\textwidth}
 \centering
 \includegraphics[height=\imheight]{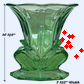}
 \includegraphics[height=\imheight]{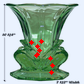}
 \includegraphics[height=\imheight]{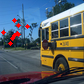}
 \includegraphics[height=\imheight]{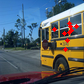}
 \includegraphics[height=\imheight]{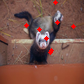}
 \includegraphics[height=\imheight]{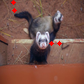}
 \\
 \includegraphics[height=\imheight]{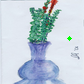}
 \includegraphics[height=\imheight]{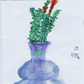}
 \includegraphics[height=\imheight]{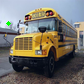}
 \includegraphics[height=\imheight]{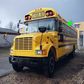}
 \includegraphics[height=\imheight]{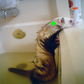}
 \includegraphics[height=\imheight]{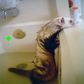}
    \caption{Visualisation of the learned dynamic sampling strategy. The red dots represent the sampled position in the support image that used to predict filter weights for the corresponding query feature position (green dot). With dynamic sampling, our model can utilize the truly informative positions to generate DMF.}
    \label{fig:visual}
 \vspace{-0.2in}
\end{figure}

\noindent\textbf{Effectiveness of dynamic sampling} 
We test models with and without dynamic sampling. As quantitative result in Tab.~\ref{tab:ablation}, the model with dynamic sampling is 1.03\% better on 1-shot and 1.50\% better on 5-shot than that without dynamic sampling. This shows that when directly using neighbor of each position to generate DMF, the model can only receive insufficient useful knowledge, leading to worse performance, which justifies the efficacy of this module. As an addition, we show the learned sampling strategy for some regions of support and query pairs in Fig.~\ref{fig:visual}. This reflects that given specific query images, the dynamic sampling can successfully learn where to collect the knowledge, thus generating more useful filters for each region.

\noindent\textbf{Do we have to delete the pooling layer?} The only difference between our backbone and the commonly-used ResNet12 is that we delete the last pooling layer, as discussed before. Tab.~\ref{tab:ablation}(b) shows the result when keeping this layer. The deletion of the pooling layer brings 0.64\% improvement on 1-shot and 1.16\% improvement on 5-shot. One reason is that when score map is larger, we can assign the loss function $\ell_f$ with more dense ground truth. Moreover, by abandoning this pooling layer, we can avoid information loss, thus helping learn better meta-filters.

\noindent\textbf{Effectiveness of group convolution} Our model is varied by different numbers of groups, as in Tab.~\ref{tab:ablation}(c). It reflects that as number of groups decreases, calculation between DMF and the corresponding region on query images gets simpler. When the number of groups equals to 1, this calculation degenerates from dot product between two vectors to element-wise product between scalar and vector, which is the same as in FEAT and CAN. This severely hurts the ability to express the whole information, leading to the worst performance. On the other hand, the profit brought by increasing number of groups saturates at different status, with 64 for 1-shot tasks and 160 for 5-shot tasks, and more groups would result in more computation consumption but along with performance drop. This means our assumption of shared information among channels in Sec.~\ref{sec:group} holds. Since support knowledge only comes from one image for 1-shot tasks, the amount of information is less than that in 5-shot tasks. Hence, 64 filters each position are sufficient for handle 1-shot tasks, while the number of filters needs to be increased to 160 for 5-shot tasks to deal with more information.

\begin{figure}
    \centering
    \includegraphics[scale=0.2]{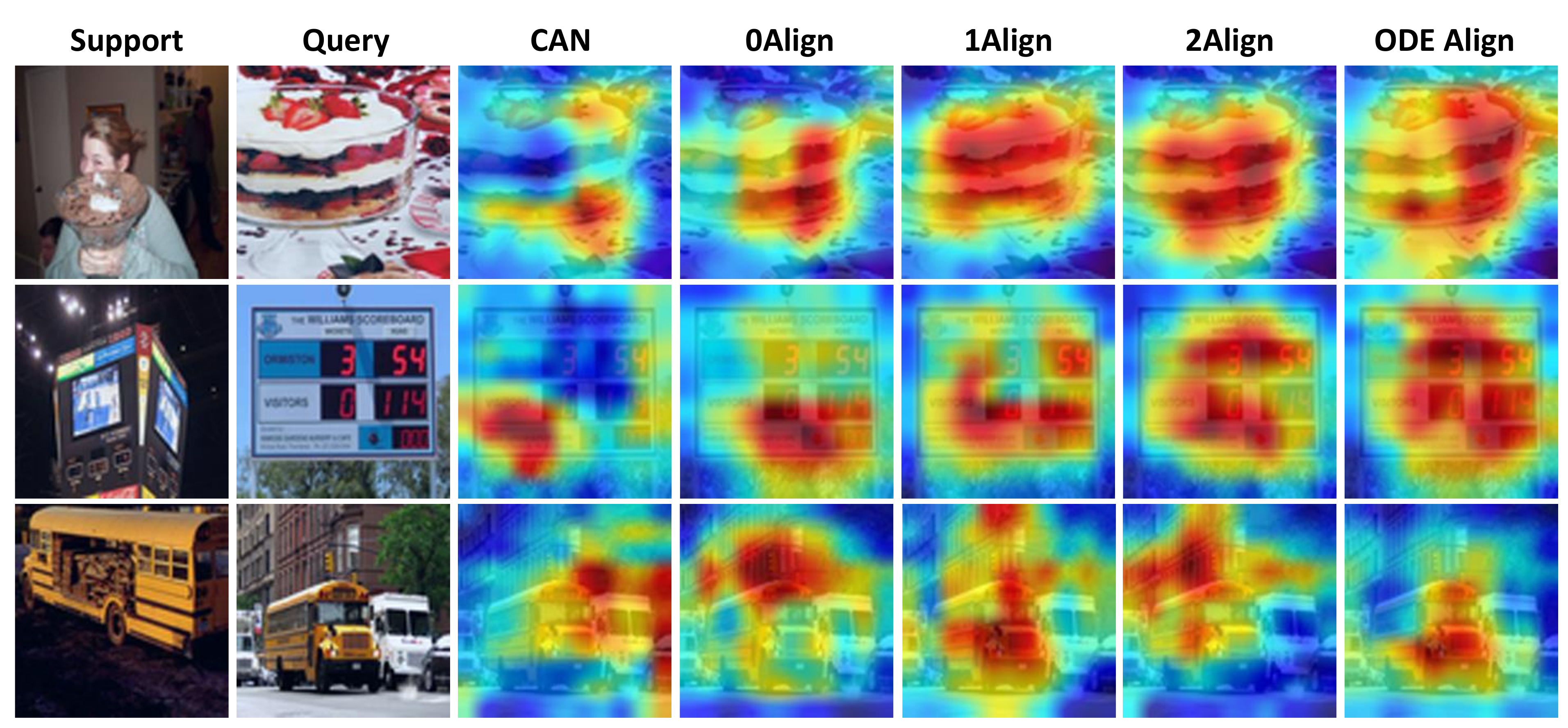}
    \caption{Comparison of aligned query feature maps from different methods for 5-way 1-shot tasks on \textit{mini}ImageNet.}
    \label{fig:feat_vis}
    \vspace{-0.2in}
\end{figure}

\noindent\textbf{Instantiation} We compare four different kinds of instantiations of our propose methods as follows: (1) 0 align: The whole model is used without DMF, where the support and query features extracted with backbone network are directly delivered to meta-classifier. (2,3) 1/2 align: The DMF is used with fixed number of recursion. (4) ODE align: Neural ODE is used in place of recursive alignment, which is our final model. The results in Tab.~\ref{tab:ablation}(d) show that using one alignment can improve the model with no alignment by 1.28\% on 1-shot and 0.77\% on 5-shot, which verifies the motivation of using dynamic meta-filter for feature alignment. Furthermore, an interesting fact is that while using more alignments can slightly improve the performance on 1-shot tasks, the 5-shot accuracy dramatically decreases, with a 0.95\% gap between models with 1 and 3 alignments. This phenomenon reflects the inflexibility of fixed number recursive alignment when dealing with different types of tasks. Compared to this, using ODE align can boost the performance on both 1-shot and 5-shot tasks. Similar results are shown in Fig.~\ref{fig:feat_vis}, where we present three pairs of support and query images and the corresponding query feature maps from the above instantiations of our model along with CAN. We find that (1) With our dynamic alignment, the results are generally better than the ones generated by CAN and model with no alignment. Our DMF is able to focus on the target object, which suggests the efficacy of our method. (2) The model with one and two alignments can only handle different data. While both models can correctly align the image in the first row, the model trained with two alignments can have better representation than that with one alignment in the second row, and it is opposite in the third row. In contrast, the ODE align can correctly highlight the query feature in all circumstances, which proves our claim of the requirement of adaptive alignment.

\subsection{Further Discussion}
Our proposed dynamic meta-filter works well with the adaptive alignment in several few-shot learning datasets as shown. However, it cannot be ignored that some extreme conditions would weaken our method. For example, when the target object appears in very different parts of support and query images, e.g., upper left and lower right, it would be hard for our model to learn such an alignment. Even though such a case is too extreme to hurt the performance, finding a remedy for that will be a potential future work.

\section{Conclusion}
We propose to learn a novel class recognition network for few-shot learning with a novel dynamic meta-filter generated from the few-shot inputs.
We dynamically sample a relevant neighbour for each feature position of few shot input and further predict position-specific and channel-specific filter weights based on the sampled neighbourhood to facilitate novel class recognition. 
This formulation is able to better capture position based semantic context of the few-shot example and thus enjoys better dynamical knowledge adaptation for few-shot learning.
This is demonstrated by the fact that we establish new state-of-the-arts on major benchmarks \textit{mini}ImageNet and \textit{tiered}ImageNet .

{\small
\bibliographystyle{ieee_fullname}
\bibliography{egbib}
}

\end{document}